\documentclass[conference,a4paper]{IEEEtran}
\IEEEoverridecommandlockouts 
\usepackage{cite}
\usepackage{amsmath,amssymb,amsfonts}
\usepackage{algorithmic}
\usepackage{graphicx}
\usepackage{eso-pic}
\usepackage{float} 
\usepackage{textcomp}
\usepackage{xcolor}
\usepackage{url}
\usepackage{booktabs}
\usepackage{array}
\usepackage[hidelinks]{hyperref}
\hypersetup{
  pdftitle={Passing Tests, Missing Semantics: Evidence of Interpretive Lock-In in Plan-Based DataFrame Code Generation},
  pdfauthor={Jingjing (May) Liu; Dishu Yang; Hongbo Qin; Hansong Zhang}
}

\makeatletter
\def\ps@IEEEtitlepagestyle{%
  \def\@oddfoot{\mycopyrightnotice}%
  \def\@evenfoot{}%
}
\def\mycopyrightnotice{%
  {\footnotesize 979-8-3195-3450-7/26/\$31.00~\copyright~2026 IEEE\hfill}%
  \gdef\mycopyrightnotice{}%
}
\makeatother

\newcommand\AtPageUpperMyright[1]{\AtPageUpperLeft{%
  \put(\LenToUnit{0.17\paperwidth},\LenToUnit{-2cm}){%
    \parbox{0.9\textwidth}{\raggedleft\fontsize{8}{11}\selectfont #1}}}}
\newcommand{\conf}[1]{\AddToShipoutPictureBG*{\AtPageUpperMyright{#1}}}

\def\BibTeX{{\rm B\kern-.05em{\sc i\kern-.025em b}\kern-.08em
    T\kern-.1667em\lower.7ex\hbox{E}\kern-.125emX}}

\begin{document}

\title{\vspace*{1cm}Passing Tests, Missing Semantics: Evidence of Interpretive
Lock-In in Plan-Based DataFrame Code Generation}

\author{
\IEEEauthorblockN{Jingjing (May) Liu\textsuperscript{*,\textdagger}}
\IEEEauthorblockA{\textit{University of California,} \\
\textit{Berkeley} \\
Berkeley, CA, USA \\
mayliujj@berkeley.edu \\
\textsuperscript{\textdagger}Corresponding author}
\and
\IEEEauthorblockN{Dishu Yang\textsuperscript{*}}
\IEEEauthorblockA{\textit{Northeastern University} \\
San Jose, CA, USA \\
yang.dis@northeastern.edu}
\and
\IEEEauthorblockN{Hongbo Qin}
\IEEEauthorblockA{\textit{Northeastern University} \\
Boston, MA, USA \\
qin.ho@northeastern.edu}
\and
\IEEEauthorblockN{Hansong Zhang}
\IEEEauthorblockA{
\textit{University of California,} \\
\textit{Berkeley} \\
Berkeley, CA, USA \\
hansong@berkeley.edu}
\thanks{\textsuperscript{*}Equal contribution as co-first authors.}
}

\IEEEaftertitletext{\vspace{-6pt}}
\maketitle
\conf{\textit{Proc. of International Conference on Electrical and Computer
Engineering Researches (ICECER 2026) \\
3-5 December 2026, Istanbul - T\"urkiye}}

\begin{abstract}
Planning and revision are intended to bring generated code closer to the task
specification. Yet a mistaken interpretation can persist across the plan,
implementation, and tests---a pattern we call interpretive lock-in. We investigate
this risk on 30 DataFrame tasks derived from TPC Benchmark DS, comparing Frozen
One-Pass, Test-Guided Revision, and PlanAudit-Guided Revision. Test-pass counts
rose from 22/30 to 29/30 and 30/30, yet no generated solution matched its stored
reference output. Evaluator controls returned the expected outcome in 57/57
cases. Same-engine SQL re-execution reproduced 28/30 references; all 77 test-passing
runs on these tasks still failed reference agreement. PlanAudit-Guided
Revision improved code scores on 3/30 tasks and test coverage on 11/30, but the
median paired change was zero for both measures. Failure analysis revealed tests that omitted
requirements, adopted candidate-defined schemas, used nondiscriminating fixtures,
or repeated the implementation's incorrect logic. Fresh generation on ten
targeted tasks produced 0/30 reference matches and 25/25 false passes among
test-passing runs. These findings are diagnostic rather than causal: the pipelines
differ in feedback, grounding, call budgets, and editing permissions, and the
study uses one model and one primary generation per eligible task.
Because pre-adjudication human labels were not retained, inter-rater agreement
cannot be reconstructed and the mechanism analyses remain exploratory.
Overall, increasingly successful
self-generated tests did not establish semantic correctness, even after excluding
unreproduced references.
\end{abstract}

\begin{IEEEkeywords}
interpretive lock-in, code generation, generated tests, semantic verification,
DataFrame transformations
\end{IEEEkeywords}

\section{Introduction}
\label{sec:introduction}

Plan-based code generation separates planning from implementation, but the
plan, program, and tests may share one incomplete interpretation. This risk is
acute for DataFrame transformations, whose behavior depends on table choice,
joins, filters, grouping grain, metrics, schema, and ordering. EvalPlus exposed
incorrect generated programs that passed the original benchmark tests by
expanding the test suites~\cite{b4}. We examine how missed requirements persist
across plans, implementations, and self-generated tests.

Let $I$ be the interpretation shared by the plan, implementation, and tests.
Passing tests establish agreement with assertions under $I$, not that $I$
matches the task. We call the persistence of shared omissions or mistaken
requirements through revision \emph{interpretive lock-in}. Unlike a false pass,
this artifact-level pattern requires evidence that the same mistake survives
revision; it does not describe internal model cognition.

We ask: \textbf{RQ1}, how do three complete pipelines differ in unordered
TPC-DS SF=1 reference agreement; \textbf{RQ2}, how do plan, code, and test
obligation-coverage outcomes change within tasks; and \textbf{RQ3}, how often
do tests pass despite reference mismatch?

\textsc{PlanAudit-DF} extracts prompt-derived obligations, audits their coverage
in a plan, PySpark program, and tests, and returns revision feedback; hidden
reference outputs enter only post-hoc evaluation. We compare Frozen One-Pass,
Test-Guided Revision, and PlanAudit-Guided Revision on TPC-DS-derived
tasks. Contributions are: (1) semantic-obligation retention as a
plan-mode measure; (2) a prompt-derived audit pipeline; (3) a 30-task evaluation
set with hidden references and obligation labels; and (4) a paired, single-realization
comparison of three complete pipelines. We supplement that comparison with
positive and negative evaluator controls, SQL re-execution, and sensitivity
analyses that separate the observed verification gap from unresolved reference
provenance. On the 28 tasks whose stored references were reproduced through
SQL re-execution, all 77 test-passing suites remained reference mismatches.
\section{Related Work}
\label{sec:related-work}

\textbf{Planning and intent alignment.} Self-planning separates solution
planning from implementation~\cite{b1}. TiCoder clarifies user intent through
interactive judgments on generated tests~\cite{b3}. Our study instead tracks
whether requirements survive a model-generated plan, implementation, and test
suite without user clarification during revision.

\textbf{Generated-test reliability.} Chen et al. show that biased self-generated
tests can mislead post-execution debugging~\cite{b5}. ACES weights tests by
leave-one-out agreement with candidate rankings rather than counting all test
votes equally~\cite{b18}. Code-A1 trains code and test models with opposing
objectives to address self-collusion~\cite{b19}. These approaches target repair,
selection, or training; our diagnostic analysis traces missed requirements
across artifacts and compares suite outcomes with a withheld SQL reference.

\textbf{Semantic evaluation and specifications.} DS-1000 combines execution
checks with API constraints for data-science code~\cite{b8}. For text-to-SQL,
Zhong et al. evaluate query behavior over distilled database test
suites~\cite{b16}. Clover combines formal verification and LLM-based checks of
consistency among code, docstrings, and formal annotations~\cite{b10}. VeriSpecGen
links atomic requirements to targeted tests and localized repairs of Lean
specifications~\cite{b11}. \textsc{PlanAudit-DF}
instead audits natural-language obligations in PySpark artifacts without formal
guarantees, separates prompt-inferable requirements from hidden output contracts,
and reserves SQL-reference comparison for evaluation after revision.
\section{Methodology}
\label{sec:experimental-setup}
\label{sec:methodology}

\subsection{Benchmark and Task Construction}

TPC-DS provides warehouse schemas, data generation, and analytical
SQL~\cite{tpcds4}. We execute its 4.0 reference queries with Spark 4.1.2 using
the KubedAI \texttt{spark-sql-perf} fork (branch
\texttt{support-tpcds-v4.0}, commit \texttt{6063ccd}). The fork supplies 99 numbered
templates and four a/b pairs, yielding 103 variants~\cite{kubedai}. Generated
programs use the PySpark DataFrame API~\cite{sparkdf} and expose
\texttt{solve(spark, tables) -> DataFrame}, with input DataFrames keyed by
table name. Stored outputs are the evaluation targets; reference-construction
exceptions and SQL re-execution are assessed in
Section~\ref{sec:reference-sensitivity}.

Before viewing model output, we retained variants that
(i) admit one stateless \texttt{solve} function, (ii) execute unmodified on
Spark 4.1.2, and (iii) use the permitted DataFrame API without user-defined functions (UDFs), lateral
views, or unavailable extensions. Thirty qualified and 73 were excluded,
including recursive-CTE, multi-statement, pivot, and unsupported-extension
tasks. All 30 tasks assessed as satisfying these eligibility criteria were
retained; no additional sampling was performed among eligible tasks. One author
assessed eligibility without independent double-coding.
All runs use the same SF=1 database to hold the input instance fixed across
conditions.

One author translated each retained query into \texttt{task.yaml}; another
author checked prompt fidelity. Prompts specify input tables,
joins, filters, aggregations, schema, ordering, and requested artifacts
(\texttt{plan.md}, \texttt{solution.py}, and \texttt{test\_solution.py}) without
revealing SQL. Prompts and reference Parquet outputs were frozen before
generation.

We derived 334 reference obligations (mean 11.1 per task, range 8--12) from SQL, the specification, and output schemas. \emph{Prompt-inferable}
obligations follow from \texttt{task.yaml}; \emph{schema-grounded} obligations
add input-schema metadata; \emph{hidden-reference} obligations require gold SQL
or output inspection (e.g., exact aliases, inferred decimal widths, and rollup
\texttt{NULL} semantics). Hidden-reference failures measure SQL-contract
agreement rather than prompt satisfaction. After all runs froze, both authors independently labeled coverage and failure
mechanisms under a frozen rubric using plans, code, tests, and oracle evidence.
The second author did not see the first author's labels before adjudication.
Original pre-adjudication human labels were not retained, so inter-rater
agreement cannot be reconstructed. The export pairs automated and final labels,
not independent human labels. These reference-based labels differ from the
prompt-derived audit labels below.

\subsection{Conditions and Comparative Scope}
\label{sec:experimental-conditions}

Each task starts with one generation call; both revision branches reuse its
artifacts. The budgets below count protocol-level model calls, excluding
pytest/Spark executions, not measured API requests. Grounding, feedback, and
revision permissions (Table~\ref{tab:pipelines}) define pipeline-level comparisons.

\textbf{Frozen One-Pass (one call).} Given the task and input schemas (names, types, and
nullability; no rows), the model produces all artifacts once. Tests and a
non-evaluative DataFrame smoke check run without revision. The smoke check is
separate from the primary evaluator.

\textbf{Test-Guided Revision (at most three calls).} The shared initial call
plus up to two revision calls gives this budget; each revision receives pytest
failures. The model may revise \texttt{solution.py} and
make syntax, import, or fixture additions to \texttt{test\_solution.py}, but may
not add gold-derived assertions; \texttt{plan.md} remains frozen. Observed
permission deviations are reported in Section~\ref{sec:verification-illusion}.

\textbf{PlanAudit-Guided Revision (at most eight calls).} The budget includes
the shared initial call, five audit calls, and up to two revision calls. The six
stages below are workflow steps, not separate agents; C-2.5 inserts schema
grounding between C-2 and C-3.
\begingroup
\setlength{\topsep}{2pt}
\begin{itemize}
\setlength{\itemsep}{0pt}
\setlength{\parsep}{0pt}
\item \textbf{C-1: Extract.} Derive obligations from the task prompt.
\item \textbf{C-2: Assess.} Label whether the prompt supports each obligation.
\item \textbf{C-2.5: Ground.} Resolve ambiguities using schema metadata,
without input rows.
\item \textbf{C-3: Audit.} Check obligations against the one-pass plan, code,
and tests; label coverage \texttt{covered}, \texttt{partial},
\texttt{absent}, or \texttt{contradicted}.
\item \textbf{C-4: Recommend.} Convert prior audit outputs into actionable
repair feedback.
\item \textbf{C-5: Revise.} Use feedback and artifact copies to revise all
three artifacts and rerun pytest for at most two cycles. Stop when tests pass;
skip revision if no actionable gap remains.
\end{itemize}
\endgroup
Calls share condition-specific context but receive only prior structured
outputs and the stage-specific inputs above.

\begin{table}[H] 
\caption{Artifact revision permissions.}
\label{tab:pipelines}
\label{tab:permissions}
\centering
\footnotesize
\setlength{\tabcolsep}{2pt}
\begin{tabular}{@{}lccc@{}}
\toprule
Artifact & One-Pass & Test-Guided & PlanAudit \\
\midrule
\texttt{plan.md} & Frozen & Frozen & Revise \\
\texttt{solution.py} & Frozen & Revise & Revise \\
\texttt{test\_solution.py} & Frozen & Limited$^{a}$ & Revise \\
\bottomrule
\multicolumn{4}{@{}l}{\scriptsize $^{a}$Syntax/import/fixture edits; no gold assertions.}
\end{tabular}
\end{table}

\subsection{Agent Information Isolation}
\label{sec:agent-isolation}

Each condition uses a separate directory and reset context. A preflight check
excludes reference SQL/outputs, human labels, scores, and other-condition
artifacts, except the shared one-pass snapshot. Candidate artifacts enter the
audit at C-3; revisions receive no reference-comparison results. Artifacts freeze
before external evaluation. These controls are procedural; operating-system
enforcement and access compliance were not verified.

\subsection{Execution Protocol}

Intent-to-treat assignment retains permission deviations; post-hoc diffs classify
plan, solution, and test edits.
Claude Code (July 2026) was configured with \texttt{claude-sonnet-4-6[1m]},
temperature 1.0, and top-p 1.0. The suffix requests a one-million-token context
window~\cite{claudeconfig}. The command-line interface (CLI) build, serving API, returned model ID, and
per-run invocation counts are unavailable; only call budgets are reported.
Execution uses Spark/PySpark 4.1.2, Python 3.11.15,
\texttt{local[*]}, eight shuffle partitions, and one machine. Generation uses
\texttt{SCALE=1}, \texttt{useDoubleForDecimal=false}, and
\texttt{useStringForDate=false}; retained settings include Java, adaptive
execution, ANSI, time zone, case sensitivity, partitioning, and parallelism.
Token counts were not recorded; no cost or efficiency comparison is made.

\textbf{Artifact availability.} Materials are available from the corresponding
author upon reasonable request: run and obligation-label exports, paired
analyses, evaluator and validation scripts, reference-reproduction reports,
and a source-file registry. The accompanying validation report documents
provenance, implementation checks, and remaining artifact gaps.

\subsection{Evaluation and Validation}
\label{sec:evaluation}
\label{sec:validation-method}

\textbf{Reference agreement and false passes.} Candidate and reference outputs
must have matching lower-cased column names and Spark types in declared column
order, equal row counts, and identical row multisets, including duplicates.
Requested row order, nullability flags, and field metadata are not evaluated;
reference agreement therefore does not establish compliance with requested
ordering. A \emph{false pass} occurs when pytest exits 0 with no failed, errored,
or collection-failed cases but reference agreement fails. The false-pass rate
uses all assigned runs as its denominator; the conditional false-pass proportion
uses test-passing runs.

\textbf{Obligation scores and paired analysis.} Means include all 30 task scores
per condition, each calculated from final labels over 8--12 obligations (334 per
condition) and checked against the run export. Plan weights are present=1,
partial/ambiguous=0.5, absent/contradicted=0; code weights are satisfied=1,
partial/violated/execution-failed=0; test weights are covered=1,
partially-covered=0.5, not-covered=0. Execution failures receive zero code scores;
plan and test scores retain observed labels. A self-confirming omission (SCO)
requires an absent/contradicted
plan, violated code (excluding execution failure), and an uncovered test
obligation, recomputed from labels.

For each revision branch, task-level changes subtract the inherited one-pass
score. We report improved/unchanged/decreased counts, median changes, and
quartiles as medians of the lower and upper halves (Tukey hinges). Binary
outcomes derive from execution-comparison records; scores derive from final
obligation labels. Results are descriptive, without hypothesis tests or
population-rate intervals.

\textbf{Robustness analyses.} Twenty-five tasks have no double-precision output
and use exact equality. Post-hoc comparisons for q2, q13, q17, q35, and q78 align
rows on query-specific keys documented in the validation report; duplicate keys
or unmatched keys in either direction fail. Non-double values remain exact;
null matches null, not-a-number (NaN) matches NaN, and infinities require equal
signs. Candidate doubles $x$ and reference values $y$ must satisfy
$|x-y|\leq\mathrm{atol}+\mathrm{rtol}|y|$, with equal absolute and relative
tolerances of $10^{-8}$, $10^{-6}$, $10^{-5}$, and $10^{-4}$.
A separate post-hoc ladder for PlanAudit-Guided outputs applies (0) exact
comparison, (1) positional alias normalization, (2) numeric casting to 64-bit
floating point, and (3) row sorting. Re-execution can change nondeterministic
row sets; Section~\ref{sec:normalized-comparison} reports one instance.

\textbf{Evaluator controls and reference checks.} Controls invoke the production
comparator with row-order enforcement disabled: unchanged loads of 33 references
(the 30 study queries plus q23b, q54, and q85), row permutations, and seven
rejection categories (Table~\ref{tab:evaluator-controls}). They were withheld
from generation and revision; detailed records are in the validation report.
A separate Spark session re-executes SQL on SF=1 against unchanged stored
references. Sensitivities exclude q11's rerun-derived PlanAudit score, retain
SQL-reproduced tasks, or apply both restrictions, preserving task pairing and
original primary outcomes.

\section{Results}
\label{sec:results}

The primary study comprises 90 condition--task pairs across 30 tasks. Unless
stated otherwise, counts follow Frozen One-Pass, Test-Guided Revision, and
PlanAudit-Guided Revision, using Section~\ref{sec:evaluation}'s definitions.

\subsection{Evaluator Validation}
\label{sec:evaluator-controls}

All 57 controls returned the expected outcome (Table~\ref{tab:evaluator-controls}):
36 acceptances and 21 rejections, with no skips and all 24 perturbations marked
verified. Each perturbation category was tested on three queries. Numeric and null mutations preserved schema and row count but failed
multiset comparison, exercising value checks separately. Row permutations were
accepted, consistent with unordered evaluation.

\begin{table}[!ht]
  \caption{Evaluator control results.}
  \label{tab:evaluator-controls}
  \centering
  \footnotesize
  \setlength{\tabcolsep}{3pt}
  \begin{tabular}{@{}lcc@{}}
    \toprule
    Control & Expected & Observed \\
    \midrule
    Unchanged reference & Accept & 33/33 accepted \\
    Row permutation & Accept & 3/3 accepted \\
    Column alias change & Reject & 3/3 rejected \\
    Column-order change & Reject & 3/3 rejected \\
    Row deletion & Reject & 3/3 rejected \\
    Row duplication & Reject & 3/3 rejected \\
    Decimal-precision change & Reject & 3/3 rejected \\
    Numeric-value change & Reject & 3/3 rejected \\
    Null insertion & Reject & 3/3 rejected \\
    \bottomrule
  \end{tabular}
\end{table}

\subsection{Reference Agreement and Internal Metrics}
\label{sec:zero-gold-match}
\label{sec:verification-illusion}

No candidate matched the reference (0/30 per condition), although test-pass
rates rose from 22/30 to 29/30 and 30/30
(Table~\ref{tab:condition-summary}). Every passing suite was therefore a false
pass: the conditional false-pass proportion was 100\% in each condition.
Execution succeeded in 28/30, 29/30, and 29/30 runs. All run outcomes and score means retain the 30 assigned
tasks in each condition; failed executions contribute zero code scores.

\begin{table}[!ht]
  \caption{Pipeline outcomes and mean obligation scores.}
  \label{tab:primary-results}
  \label{tab:condition-summary}
  \label{tab:oracle-failures}
  \centering
  \scriptsize
  \setlength{\tabcolsep}{3pt}
  \begin{tabular}{@{}lccc@{}}
    \toprule
    Metric & One-Pass & Test-Guided & PlanAudit \\
    \midrule
    Tests pass & 22/30 & 29/30 & 30/30 \\
    Reference match & 0/30 & 0/30 & 0/30 \\
    False pass & 22/30 & 29/30 & 30/30 \\
    False-pass rate & 73.3\% & 96.7\% & 100.0\% \\
    Mean plan score & 89.16\% & 89.16\% & 90.34\% \\
    Mean code score & 76.73\% & 79.07\% & 81.01\% \\
    Mean test coverage & 81.25\% & 84.75\% & 87.70\% \\
    \bottomrule
  \end{tabular}
\end{table}

\noindent\textbf{Within-task changes (RQ2).}
Table~\ref{tab:paired-changes} reports all 30 paired tasks per branch, with
score changes in percentage points. No score
decreased in the primary export, and all median changes were zero. Only
PlanAudit test coverage had nonzero IQR (8.33 percentage points): mean gains
reflect improvements in a subset rather than across most tasks.

\begin{table}[!ht]
  \caption{Within-task score changes from Frozen One-Pass.}
  \label{tab:paired-changes}
  \centering
  \scriptsize
  \setlength{\tabcolsep}{2.5pt}
  \begin{tabular}{@{}llcc@{}}
    \toprule
    Revision & Score & Up/Same/Down & Median [Q1, Q3] \\
    \midrule
    Test-Guided & Plan & 0/30/0 & 0 [0, 0] \\
    Test-Guided & Code & 1/29/0 & 0 [0, 0] \\
    Test-Guided & Test & 3/27/0 & 0 [0, 0] \\
    PlanAudit-Guided & Plan & 4/26/0 & 0 [0, 0] \\
    PlanAudit-Guided & Code & 3/27/0 & 0 [0, 0] \\
    PlanAudit-Guided & Test & 11/19/0 & 0 [0, 8.33] \\
    \bottomrule
  \end{tabular}
\end{table}

Excluding the session-sensitive q11 score record leaves PlanAudit gains on
3/29 plans, 2/29 code scores, and 10/29 test scores, with no decreases and zero
median changes. Its mean code gain is then 3.28 rather than 4.28 percentage
points.

For PlanAudit-Guided Revision, the primary failure categories were alias
mismatch (15), structural schema (4), row count (3), precision (3), values (3),
column order (1), and runtime (1). Alias and precision account for 18 cases
with hidden-reference contract mismatches; 11 concern prompt-inferable
requirements. These are surface categories, not mutually exclusive root causes:
normalization below tests whether additional differences survive contract
relaxation.

Test-Guided Revision gained 23.3 percentage points in test passing. Its code
mean rose 2.33 points, entirely through q2's recovery from execution failure
to 70\% obligation satisfaction. One collection-error trajectory substantially
expanded tests under fixture permissions; it remains assigned to Test-Guided
Revision and is reported as a protocol deviation. PlanAudit's 4.28-point code
gain spans q4, q11, and q2, while test coverage rose 6.45 points without recovering
primary reference agreement.

\begin{figure}[!ht]
  \centering
  \includegraphics[width=\columnwidth]{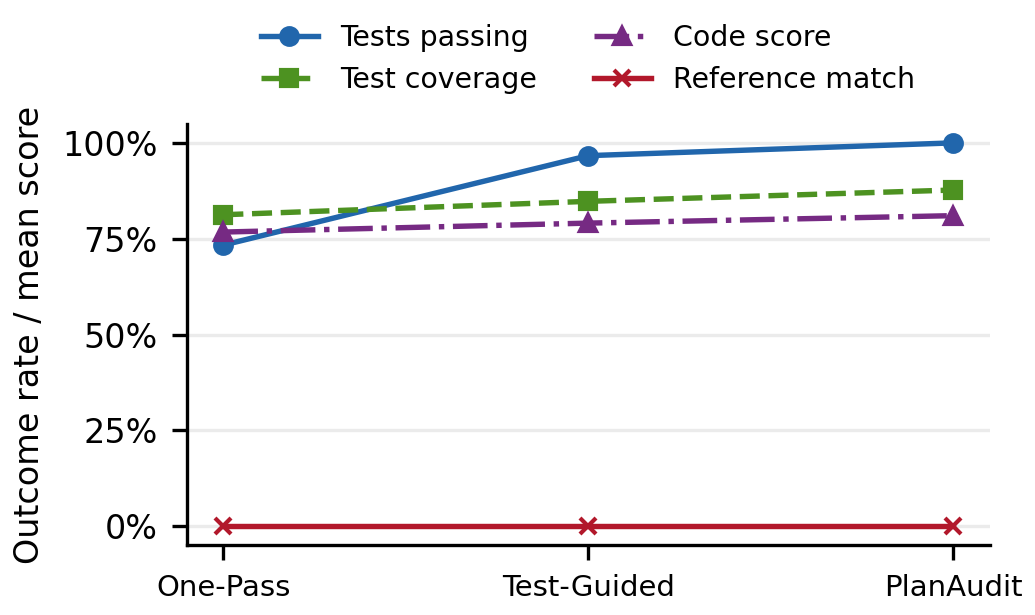}
  \caption{All-run test passing, code scores, and test coverage increase,
  while primary reference agreement remains zero ($n=30$ per condition).}
  \label{fig:verification-illusion}
\end{figure}

High internal scores did not imply reference agreement in this sample. Among
runs with plan semantic ratio at least 90\%, 17, 17, and 19 lacked agreement;
among runs with test coverage at least 90\%, the corresponding counts were 11,
12, and 17. Figure~\ref{fig:verification-illusion} summarizes the aggregate
pattern.

\subsection{SQL Re-execution and Reference Sensitivity}
\label{sec:reference-sensitivity}

All 33 SQL queries in the reference inventory executed successfully; 31 outputs
matched the stored schema and multiset. Three matching outputs (q23b, q54, and
q85) lie outside the primary study, leaving \textbf{28/30 reproduced study
references}. This includes the case-study references q4, q11, q14a, q33, q80,
and q5, as well as the alias-only recovery cases q13 and q17.

For q2, re-execution returns 2,513 rows rather than 53 and differs in names and
numeric types; renaming cannot reconcile it. For q35, both outputs contain
100 rows but schemas differ, preventing value comparison. Both references are
attributed to task-specific construction whose code is unavailable, so their
provenance remains unresolved.

The primary 30-task results remain unchanged. Restricting analysis to the 28
SQL-reproduced tasks yields 21/28, 28/28, and 28/28 passing suites and 0/28
reference matches in every condition (Table~\ref{tab:reference-sensitivity}).
All 77 passing suites remain false passes. Thus, the observed separation between
test passing and reference agreement does not depend on q2 or q35.

\begin{table}[!ht]
  \caption{Test passing across reference-sensitivity cohorts.}
  \label{tab:reference-sensitivity}
  \centering
  \footnotesize
  \setlength{\tabcolsep}{3pt}
  \begin{tabular}{@{}lccc@{}}
    \toprule
    Cohort & One-Pass & Test-Guided & PlanAudit \\
    \midrule
    All assigned & 22/30 & 29/30 & 30/30 \\
    SQL-reproduced & 21/28 & 28/28 & 28/28 \\
    Also excluding q11 & 20/27 & 27/27 & 27/27 \\
    \bottomrule
  \end{tabular}
\end{table}

On the 28-task subset, mean code scores are 82.21\%, 82.21\%, and 84.30\%,
while mean test coverage is 81.34\%, 85.09\%, and 88.25\%. Each mean uses
28 tasks and 314 underlying obligations per condition. Test-Guided Revision
has no code-score changes and improves test coverage on 3/28 tasks.
PlanAudit-Guided Revision improves 4/28 plan, 2/28 code, and 11/28 test scores;
none decreases. All paired medians remain zero, with test-coverage IQR
8.33 percentage points for PlanAudit and zero for its other dimensions.

Further excluding q11 leaves 27 tasks: both revision pipelines pass every suite
without any reference match. PlanAudit improves 3 plan, 1 code, and 10 test
scores, again with no decreases and zero median changes. Its sole remaining
code improvement is q4, contributing a mean increase of 0.93 percentage points.
Thus, the broader 4.28-point gain combines localized repair, q2's execution
recovery, and q11's session-sensitive record, while the test/reference gap
survives both restrictions.

\subsection{Failure Mechanisms}
\label{sec:test-failure-mechanisms}

The mechanism analysis covers 39 violated obligations in Frozen One-Pass false
passes and 52 in PlanAudit-Guided false passes
(Table~\ref{tab:test-failure-mechanisms}). No test covered the violated
obligation in 19/39 and 24/52 cases, respectively; the remaining 20/39 and 28/52
had tests that passed despite the defect. These totals concern different sets
of test-passing runs, not a paired count of defects introduced by revision.

\begin{table}[!ht]
  \caption{Failure mechanisms among violated obligations.}
  \label{tab:test-failure-mechanisms}
  \centering
  \scriptsize
  \setlength{\tabcolsep}{3pt}
  \begin{tabular}{@{}lcc@{}}
    \toprule
    Mechanism & One-Pass & PlanAudit \\
    \midrule
    Gap despite plan & 13 (33\%) & 17 (33\%) \\
    Partial coverage & 10 (26\%) & 13 (25\%) \\
    Candidate-schema validation & 9 (23\%) & 11 (21\%) \\
    SCO: absent plan and test & 6 (15\%) & 7 (13\%) \\
    Nondiscriminating data & 0 & 3 (6\%) \\
    Same incorrect logic & 1 (3\%) & 1 (2\%) \\
    \bottomrule
  \end{tabular}
\end{table}

Missing checks include planned but untested obligations and SCOs spanning all
three artifacts. Inadequate checks include coarse aggregates, candidate-defined
schemas, nondiscriminating fixtures, and assertions repeating incorrect logic.

Four primary-study cases illustrate these mechanisms. In q14a, tests required
\texttt{total\_sales}, whereas the hidden contract used \texttt{sum(sales)}.
In q4, all 10 PlanAudit-revised tests passed despite using \texttt{ss\_list\_price} rather
than \texttt{ss\_ext\_list\_price}. In q33, all 11 tests passed in each condition, although aggregation obligation
O8 was labeled absent in the plan, violated in code, and untested; the output
also differed in decimal precision (17 rather than 27, both at scale 2).
For q11, the baseline labels identify an untested cross-channel customer-alignment
requirement (O6), alongside schema and ordering violations.

\subsection{Normalization and Fresh-Generation Replication}
\label{sec:normalized-comparison}

Normalization recovered only q13 and q17 as genuine alias-only mismatches;
q11's apparent recovery reflects run drift (94 rows on re-execution versus
100 originally), all three precision cases retained value differences after
casting, and 27/30 outputs failed full normalization.

A targeted replication generated fresh artifact triples for ten tasks selected
after primary analysis: three cases above and seven probing structural gaps,
SCOs, and joins (Table~\ref{tab:fresh-generation-replication}). The model,
pipelines, environment, and reference-withholding protocol were unchanged.
Fresh-generation denotes new initial artifacts, not replay of a recorded seed;
both revision branches inherit each task's new snapshot. Table entries give
passed/total tests; P, F, and E denote suite pass, failure, and execution error.
No reference matches occurred in any condition for any query (0/30 overall).

\begin{table}[!ht]
  \caption{Targeted fresh-generation replication.}
  \label{tab:fresh-generation-replication}
  \centering
  \scriptsize
  \setlength{\tabcolsep}{2pt}
  \begin{tabular}{@{}lccc@{}}
    \toprule
    Query & One-Pass & Test-Guided & PlanAudit \\
    \midrule
    q14a & 0/15 E & 15/15 P & 15/15 P \\
    q4   & 9/13 F & 9/13 F & 13/13 P \\
    q33  & 10/10 P & 10/10 P & 10/10 P \\
    q11  & 10/10 P & 10/10 P & 10/10 P \\
    q77  & 12/12 P & 12/12 P & 12/12 P \\
    q80  & E & 18/18 P & E \\
    q14b & 9/9 P & 9/9 P & 9/9 P \\
    q69  & 14/14 P & 14/14 P & 14/14 P \\
    q47  & 16/16 P & 16/16 P & 16/16 P \\
    q5   & 20/20 P & 20/20 P & 21/21 P \\
    \bottomrule
  \end{tabular}
\end{table}

All test-passing runs were false passes (25/25); the other five also failed the
reference. We report this curated sample descriptively, without a population
interval. In q80, Test-Guided Revision converted execution errors into 18/18 passing tests
that preserved \texttt{"catalog"} rather than \texttt{"catalog channel"} and
incorrect rollup \texttt{NULL} semantics. q14a repeated the alias
self-confirmation, q33 the precision failure, and q11 incorrect join semantics.
In q4, PlanAudit
repaired a join fixture but retained the wrong price formula; small fixture
values made the alternatives yield equal relative ratios.

Structural errors also survived complete suites. q77 collapsed per-call-center
catalog aggregation to one row and used string rather than integer surrogate
keys; q14b returned 8 rather than 12 \texttt{ty\_}/\texttt{ly\_} columns. q69
returned one \texttt{cnt} instead of \texttt{cnt1}/\texttt{cnt2}/\texttt{cnt3}.
q47 used \texttt{ss\_ext\_sales\_price} instead of
\texttt{ss\_sales\_price}, inflating monthly sales about 50-fold. q5 combined
\texttt{net\_profit} in place of \texttt{ext\_sales\_price}, an incorrect
profit formula, decimal-width and catalog-ID spacing errors. Its PlanAudit
revision partially corrected returns but changed an outer join to an inner join,
reducing code semantic ratio from 50\% to 40\% while all 21 tests passed.

\section{Discussion}
\label{sec:discussion}

\subsection{Interpretive Lock-In Across Revisions}

Revision improved executability and obligation retention without reference
agreement; zero median changes locate gains in a minority of tasks. In the
targeted replication, PlanAudit-Guided Revision repaired q4's fixture while
retaining the wrong price formula; Test-Guided Revision repaired q80's execution
errors while retaining channel/rollup mismatches. These trajectories are consistent
with interpretive lock-in. In q5, revision introduced a join error and reduced the
code score despite passing tests. Evaluation should therefore distinguish
persistent mistakes, repairs, new defects, and changes to the assertions judging them.

Both sensitivity restrictions preserve the test/reference gap despite smaller
code gains. Accordingly, the central verification finding does not depend on
the two references with unresolved construction provenance.

\subsection{Verification Implications}

As a practical first step, we propose deriving acceptance assertions from explicit
task contracts before candidate inspection and holding them fixed during revision.
Linking each assertion to a requirement exposes missing checks; discriminating
fixtures address existing but ineffective checks. Separate generation alone does not ensure independence when specifications or
expected values share an ambiguity.

Contracts should distinguish stated from hidden requirements and separate schema,
null behavior, grouping, channel labels, and ordering from value assertions.
Fixtures should discriminate alternatives: unmatched keys for joins, unequal price
fields for metric substitutions, and multiple groups for aggregation grain.
Targeted mutations can check this discrimination; hidden tests, differential
execution, and metamorphic relations offer complementary evidence.

\subsection{Scope and Validity}

The 57 controls test comparator behavior; SQL re-execution supports repeatability,
not independent semantic correctness. Shared SQL, translation, or Spark errors
require independent implementations or hand-checked assertions to resolve.
Feedback, grounding, budgets, and editing permissions vary together, precluding
an isolated audit effect. Unretained pre-adjudication labels prevent reconstruction
of human agreement; obligation and SCO analyses remain exploratory. Without
reference-positive generated candidates, checker-positive controls do not establish
generated-test sensitivity, discrimination, or the necessity of high coverage.

One model, one primary generation per eligible task, one database, and one
environment limit generality. Eligibility was not independently double-coded;
the targeted replication establishes recurrence, not population robustness.

\subsection{Next Comparisons}
\label{sec:next-comparisons}

Follow-up studies should cross feedback, grounding, and editing permissions at
matched call budgets, using fixed assertions and independent validators.
Prospective sampling across models, repeated generations, and databases should
include reference-positive candidates. Preregistered obligation rubrics and
retained pre-adjudication labels would permit human-agreement assessment and
help distinguish mechanism effects from benchmark and annotation variation.

\subsection{Conclusion}

Higher test passing and obligation retention did not recover reference agreement,
including on the 28-task and 27-task sensitivity subsets. \textbf{Self-consistency
was not semantic validation}. Acceptance of generated code should require evidence
tied to the task contract, not only assertions that repeat the candidate's
interpretation.


\end{document}